\documentclass[10pt,twocolumn,letterpaper]{article}

\usepackage{iccv}
\usepackage{times}
\usepackage{epsfig}
\usepackage{graphicx}
\usepackage{amsmath}
\usepackage{amssymb}

\usepackage{multirow}
\usepackage{subcaption}
\usepackage{pifont}
\usepackage{float}
\usepackage{authblk}
\usepackage{adjustbox}
\usepackage[breaklinks=true,bookmarks=false]{hyperref}

\iccvfinalcopy % *** Uncomment this line for the final submission

\def\iccvPaperID{5684} % *** Enter the ICCV Paper ID here

\ificcvfinal\fi

\begin{document}

%%%%%%%%% TITLE
\title{POD: Practical Object Detection with Scale-Sensitive Network}

\author[1,2,3]{Junran Peng}
\author[2]{Ming Sun}
\author[1,3]{Zhaoxiang Zhang\thanks{Corresponding author.}}
\author[1,3]{Tieniu Tan}
\author[2]{Junjie Yan}
\affil[1]{University of Chinese Academy of Sciences}
\affil[2]{SenseTime Group Limited}
\affil[3]{Center for Research on Intelligent Perception and Computing, CASIA}

\maketitle
\ificcvfinal\thispagestyle{empty}\fi

%%%%%%%%% ABSTRACT
\begin{abstract}
Scale-sensitive object detection remains a challenging task, 
where most of the existing methods could not learn it explicitly and are not robust to scale variance.
In addition, the most existing methods are less efficient during training or slow during inference, which are not friendly to real-time applications. 
In this paper, we propose a practical object detection method with scale-sensitive network.

%%% scale 
Our method first predicts a global continuous scale , which is shared by all position, for each convolution filter of each network stage.
%%and \textcolor{red}{the global scales predicted are extremely robust across samples in our way}. 
%
To effectively learn the scale, we average the spatial features and distill the scale from channels.
%%\textcolor{red}{}
%%% practical 
%Moreover,  We argue that a practical object detection should be efficient during training and fast during inference respectively except the high performance.
%%% fix
For fast-deployment, 
we propose a scale decomposition method that transfers the robust fractional scale into combination of fixed integral scales for each convolution filter, which exploits the dilated convolution.
We demonstrate it on one-stage and two-stage algorithms under different configurations. 
% Our method get the remarkable results on both one-stage and two-stage detectors based on resnet-50 and resnet-101 backbone while adding no parameter and computation at all. 
For practical applications, training of our method is of efficiency and simplicity which gets rid of complex data sampling or optimize strategy.
During testing, the proposed method requires no extra operation and is very supportive of hardware acceleration like TensorRT and TVM. 
On the COCO {\em test-dev}, our model could achieve a 41.5 mAP on one-stage detector and 42.1 mAP on two-stage detectors based on ResNet-101, outperforming baselines by 2.4 and 2.1 respectively without extra FLOPS.
\end{abstract}

%%%%%%%%% BODY TEXT
\section{Introduction}
With  the  blooming  development of CNNs, great progresses have been achieved in the field of object detection.
% category &real application
A lot of CNN based detectors are proposed, 
% categorized into one-stage detectors and two-stage detectors, 
and they are widely used in real world applications like face detection, pedestrian and vehicle detection in self-driving and augmented reality.
%%##################################################################################################
\begin{figure}[H]
\includegraphics[width=1\linewidth]{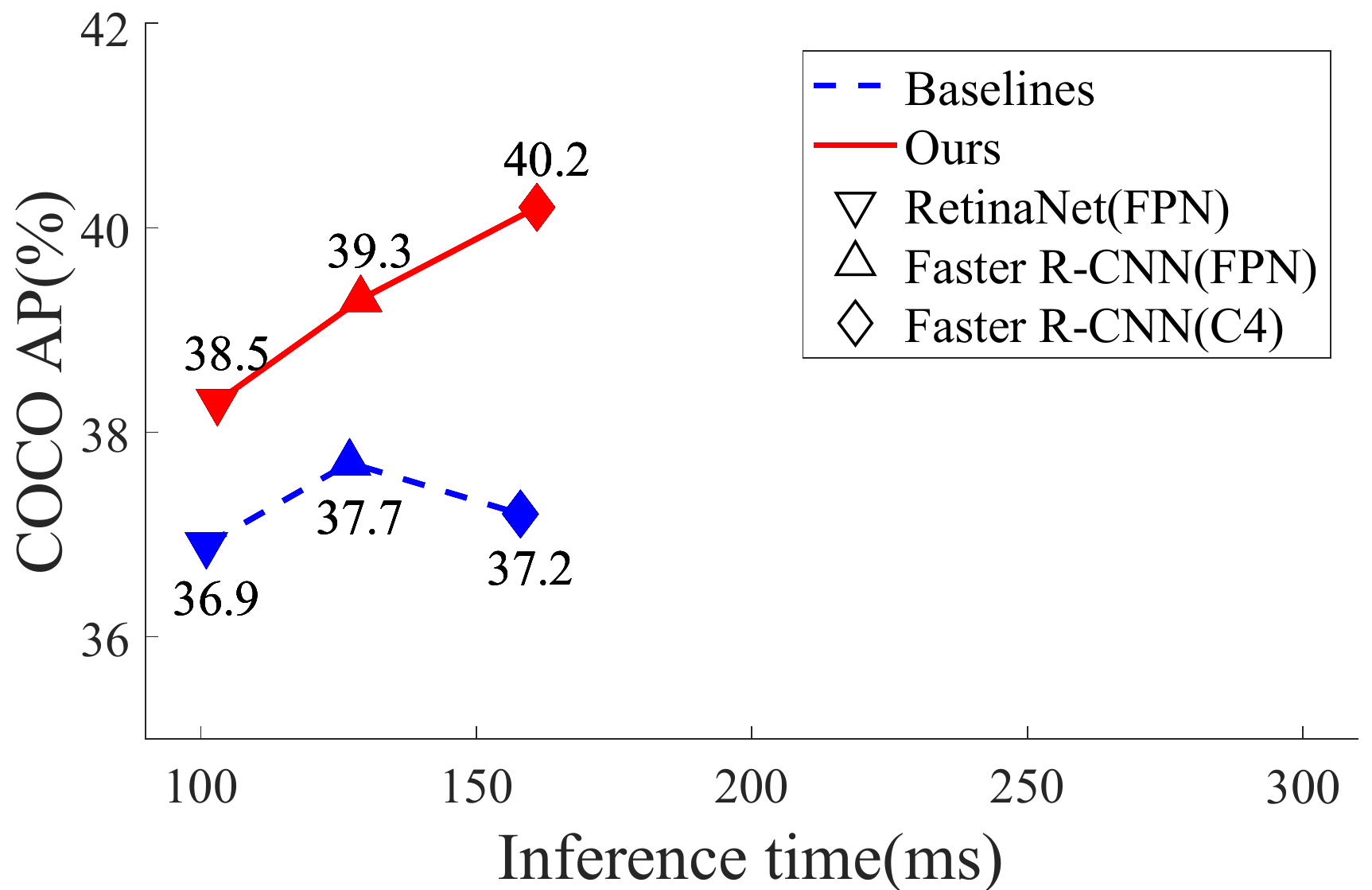}
% \hspace{-44mm}\resizebox{.51\columnwidth}{!}{
% \begin{tabular}[b]{l|cc}
% method & AP & time \\
%  \hline
%  \bd{FRCNN-C4-R50} & 37.2 & 158ms \\
%  \bf{FRCNN-C4-R50-FD} & \bf{40.2} & 161ms \\
%  \hline
%  \bd{FRCNN-FPN-R50} & 37.7 & 120ms \\
%  \bf{FRCNN-FPN-R50-FD} & \bf{39.3} & 121ms \\
%  \hline
%  \bd{RetinaNet-R50}  & 36.9 & 100ms  \\
%  \bf{RetinaNet-R50-FD}  & \bf{38.5} & 101ms  \vspace{15mm}\\
% \end{tabular} 
% }%
\caption{Speed (ms) versus accuracy (AP) on COCO {\em test-dev}. Preset with the learnt scales, our fast-deployment(FD) detectors outperforms baselines of one-stage and two-stage detectors by a large margin almost without any extra inference time.  
All of the models(R50) above are trained with $2\times$ lr schedule.
Better results and details are given in \S\ref{sec:exps}.}
\label{fig:speed}
\end{figure}
%##################################################################################################
In practical scenarios mentioned above, detection frameworks always work on mobile or embedded devices, and are strictly required to be fast and accurate. While many complicated networks and frameworks are designed to achieve higher accuracy, lots of extra parameters are imported, and larger memory cost and more inference are needed. In this paper we propose a method that could greatly improve the accuracy of detection frameworks without any extra parameters, memory or time cost.

%Most of these effective detectors rely on weights pretrained on ImageNet to such a great extent that they have to adopt network structure specially designed for image classification, otherwise the accuracy would fall rapidly training from scratch. 

%say the scale research related work
Scale variation is one of the most challenging problems in object detection.
%
% To handle scale this issue, 
% %FPN 
% many precious detectors construct feature pyramids or multi-level feature towers~\cite{lin2017feature, shrivastava2016beyond, liu2016ssd}. And multiple scale levels of
% feature maps are generating predictions in parallel.
% %
% Besides, anchor boxes can further handle scale variation~\cite{ren2015faster}
% anchors
In~\cite{ren2015faster}, anchor boxes of different sizes are predefined to handle scale variation.
% det nets/ rpn nets
Besides, some methods~\cite{Li_2018_ECCV, liu2018receptive} apply astrous convolutions to enlarge receptive field size, which makes information in larger scales captured.
% scale module such deform
In addition, some related works explicitly plug the scale module into backbone.
% excellent case 
~\cite{dai2017deformable} designs a deformable convolution to learn the scale of object to some degree, which is of local and dense learning style. 

%problem 
However, these methods contain two limitations: 1)  many predefined rules based on heuristic knowledge 2) algorithms are hostile to hardware and slow during inference.
Many object detection methods are being hampered by these constraints, especially the slow inference on mobile or embedded devices.
%
% For feature pyramids, the number of levels and rule of anchors matching are predefined, which are heuristic-guided. 
Methods like~\cite{Li_2018_ECCV} and ~\cite{liu2018receptive} are heavily heuristic-guided and could partly handle scale variation. 
While the offsets are learned efficiently in deformable convolution~\cite{dai2017deformable}, the inference is time-consuming due to the bilinear operation at each position. 
%
% The local and dense continuous offsets are hard to optimize and we show that this is not necessary.
%
% The time can be shown by ~\ref{fig:speed}, the proposed method is equal to the baseline. 
%
% While methods like DCN and SAC are effective in learning a desirable scale, 
%
In addition, the grid sampling location in convolution of these methods~\cite{dai2017deformable, zhang2017scale} are dynamic and local, which is not practical in real-time applications for the reason that fixed and integral grid sampling locations are essential for hardware optimization. 
For instance, a normal convolution network with fixed grid sampling locations could be accelerated on various kinds of devices after optimized by TVM
% \footnote{TVM is an open deep learning compiler stack for CPUs, GPUs, and specialized accelerators. See https://github.com/dmlc/tvm}
% \footnote{TVM is an open deep learning compiler stack for CPUs, GPUs, and specialized accelerators. It aims to close the gap between the productivity-focused deep learning frameworks, and the performance or efficiency-oriented hardware backends. Essentially speaking, convolutions like DCN~\cite{dai2017deformable} or SAC~\cite{zhang2017scale} are not supported in TVM due to their sampling mechanism.}
, while convolutions like DCN~\cite{dai2017deformable} or SAC~\cite{zhang2017scale} are not supported in TVM due to their dynamic and local sampling mechanisms.
In this paper, we 
%argue 
show that the local and dense continuous scales which are hard to optimize are not necessary and through collaboration of well-learnt global scales on layers, a network could be granted the scale-awareness.
Therefore, we first train a global scale learner(GSL) network to learn the distribution of continuous scale rates in $h$ and $w$ directions for layers.
%we care through GSL module.
%
The scales learnt by our GSL modules are extremely robust across samples and could be treat as fixed values which is revealed in ~\ref{sec:dist-dil}.
Secondly, the learnt arbitrary scales are transformed into combination of fixed integral dilations through our scale decomposition method.
% , which are easy to be applied on standard convolution. 
%
Once completing the decomposition, we can transfer the weights and finetune a fast-deployment(FD) network using the decomposed network architecture. 
% scale-sensitive
The blocks in FD network are granted the desirable combination of dilations which are learnt by GSL network, and the whole network could better cope with objects of a large range of scale while keeping friendly to hardware optimization. 

%experiment 
Extensive experiments on the COCO show the effectiveness of our method. 
% effectiveness of learnt dilations
With dilations learnt and weights transferred from GSL network, our FD network could outperform Faster R-CNN with C4-ResNet-50 backbone by 3.0\% AP and RetinaNet with ResNet-50 by 1.6\% AP as shown in Figure~\ref{fig:speed}. 
% effectiveness of finetuning from GSL 
% Utilizing GSL network as pretrained models, the performance could be further improved. 
Applying our method on two-stage detector based on ResNet101 could achieve 42.1\% AP on COCO {\em test-dev} with a total $2\times$ the training epochs, surpassing its $2\times$ baseline by 2.5\% AP. While on RetinaNet based on ResNet-101, our method could yield a 2.4\% AP improvement and achieve a 41.5\% AP.

Overall, there are several advantages about our method:
% The main contributions of our work are listed as follows:

\noindent - \textbf{Large Receptive Fields : } We design a module to learn a stable global scale for each layer and prove that these learnt scales could collaborate to significantly help network handle objects of a large range of scales.
% A variety of backbones designed for other tasks {\em i.e.} image classification, can be transferred to fit task of object detection through the adjustment method proposed in this paper. 

\noindent - \textbf{Fast Deployment : }
We propose a dilation decomposition method to transfer fractional scale into combination of integral dilations. The decomposed dilations which are integral and fixed enable our fast-deployment network to be fast and supportive of hardware optimization during inference.
% The adjusted model does not need to be pretrained {\em i.e.}, in ImageNet, and can make use of weights of the models pretrained in other tasks. 

\noindent - \textbf{Higher Performance : } 
Our method is widely applicable across different detection frameworks(both one-stage and two-stage) and network architectures, and could bring significant and consistent gains on both AP50 and AP without being time-consuming. We are confident that our method would be useful in practical scenarios.

\begin{figure*}[t]
\centering
\includegraphics[height=8.5cm]{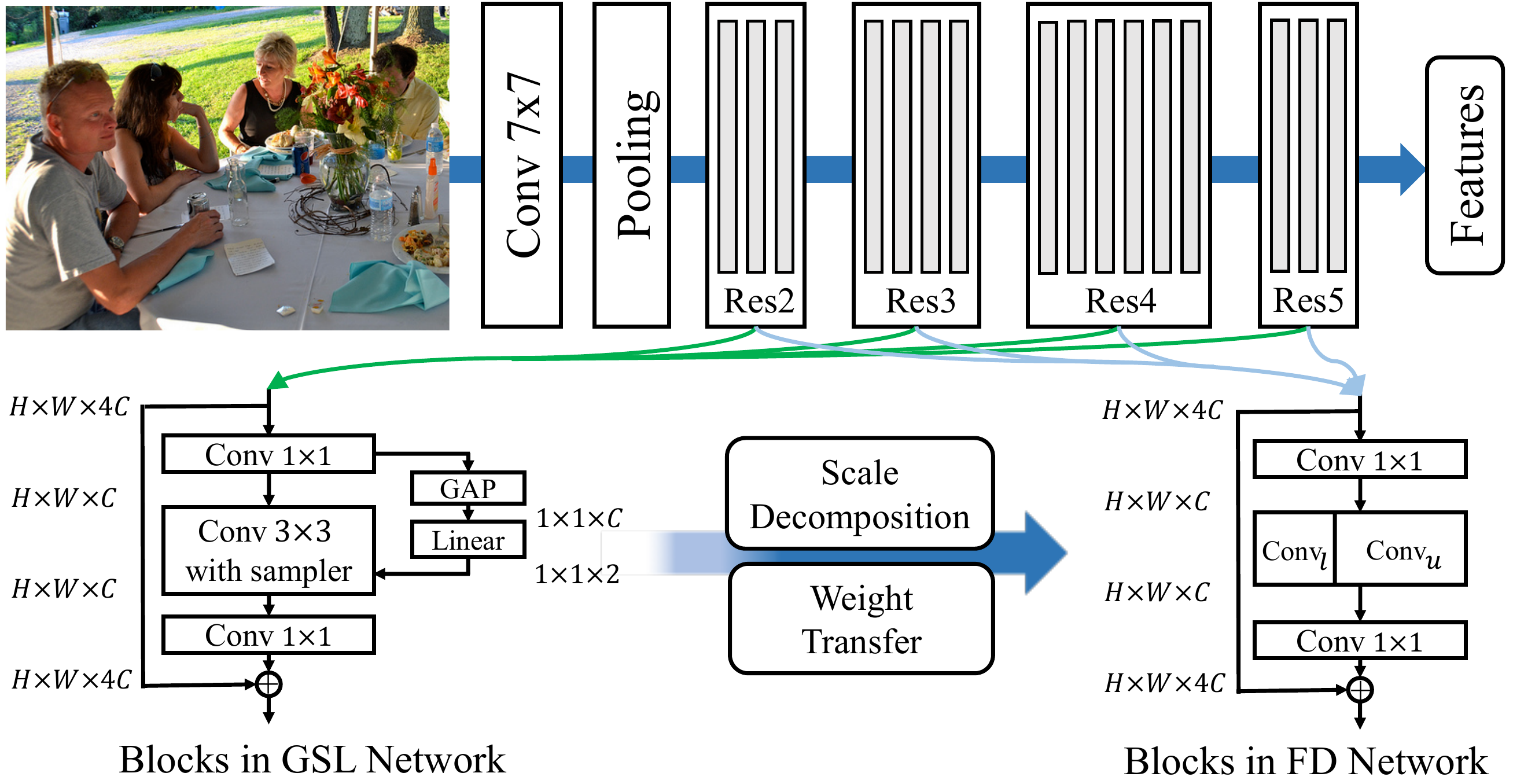}
\caption{{\bf Overview of the pipeline of our method.} We take ResNet-50 as backbone of detector in this example. We first train a GSL Network to learn a stable scale for each block in $h$ and $w$ directions respectively. The learnt scales are decomposed into combination of integral dilations for each block we care. Given groups of integral dilations, we construct a fast-deployment network and finetune it from GSL network. 
% The final FD network could be scale-sensitive and fast in practical scenario.
}
\label{fig:pipeline}
\end{figure*}

\section{Related Works }
\label{sec:related}
%------------------------------------------------------------------------
\subsection{Deep Learning for Object detection}
Object detection is one of the most important and fundamental problems in computer vision area.
% In the past years, with the blooming developments of deep learning, great progresses have been achieved in the field of object detection. 
A variety of effective networks~\cite{szegedy2015going, simonyan2014very, he2016deep, huang2017densely, xie2017aggregated, hu2018squeeze, howard2017mobilenets, sandler2018mobilenetv2} designed for image classification are used as backbone to help the task of object detection. There are generally two types of detectors. 

The first type, known as two-stage detector, generates a set of region proposals and classifies each of them by a network. R-CNN~\cite{girshick2014rich} generates proposals by Selective Search~\cite{uijlings2013selective} and extracts features to an independent ConvNet. Later, SPP~\cite{he2015spatial} and Fast-RCNN~\cite{girshick2015fast} are proposed that features of different proposals could be extracted from single map through a special pooling layer to reduce computations. Faster R-CNN~\cite{ren2015faster} first proposes a unified end-to-end detectors which introduces a region proposal network(RPN) for feature extraction and proposal generation. There are also many works that extend Faster R-CNN in various aspects to achieve better performance. FPN~\cite{lin2017feature} is designed to fuse features at multiple resolutions and provide anchors specific to different scales. Recent works like Cascade R-CNN~\cite{cai2018cascade} and IoU-Net~\cite{jiang2018acquisition} are devoted to increasing the quality of proposals to fit the COCO AP metrics. 

The second type, known as one-stage detector, acts like a multi-class RPN that predicts object position and class efficiently within one stage. YOLO~\cite{redmon2016you} outputs bounding box coordinates directly from images. SSD and DSSD involve multi-layer prediction module which helps the detection of objects within different scales. RetinaNet~\cite{lin2017focal} introduces a new focal loss to cope with the foreground-background class imbalance problems. RefineDet~\cite{zhang2018single} adds an anchor refinement module and ~\cite{law2018cornernet} predicts objects as paired corner keypoints to yield better performance. 

\subsection{Receptive Fields}
% As large scale variation of object instances is one of the most challenging problem in task of object detection, it is important for neurons in different layers to have desirable receptive fields. 
The receptive field is a crucial issue in various visual tasks, as output of a neuron only responds to information within its receptive field. 
% Some approaches grant 
Dilated convolution~\cite{yu2015multi} is one of the effective solution to enlarge receptive field size, which is widely used in semantic segmentation~\cite{chen2018deeplab, zhao2017pyramid} to incorporate contextual information. In DetNet~\cite{Li_2018_ECCV} and RFBNet~\cite{liu2018receptive}, dilated convolution are used to have larger receptive field size without changing spatial resolution. In ~\cite{jeon2017active}, convolution learns a fixed deformation of kernel structure from local information.

There are other methods that could automatically learn dynamic receptive fields. In SAC~\cite{zhang2017scale}, scale-adaptive convolutions are proposed to predict a local flexible-size dilations at each position to cover objects of various sizes. ~\cite{dai2017deformable} proposes deformable convolution in which the grid sampling locations are offsets predicted with respect to the preceding feature maps. STN~\cite{jaderberg2015spatial} introduces a spatial transformer module that can actively transform a feature map by producing an appropriate transformation for each input sample. While effective in performance, these methods are slow and unfriendly to hardware optimization in real-time applications. 
The framework we present in this paper could keep scale-aware and supportive of hardware optimization at the same time. 

\section{Our Approach}
\label{sec:approach}
An overview of our method is illustrated in Figure \ref{fig:pipeline}. 
There are mainly two steps in our approach: 
%%1) We train a dilation learner consists of dynamic-global-dilation(\textbf{dGD}) blocks
1) We design a global scale learning(GSL) module. Plain blocks are replaced with GSL modules during training to learn a recommended global scale.
% Then we randomly pick a bunch of training samples and calculate the mean values of predicted dilation of each dynamic-GD block as learnt dilations. 
2) Then we transfer the learnt arbitrary scale into fixed integral dilations using scale decomposition. 
Equipped with the decomposed integral dilations, a fast-deployment(FD) model is trained with weights transferred from GSL network. 
In this way we can acquire a high-performance model comprising of convolutions with only fixed and integral dilations, which is friendly to hardware acceleration and fast during inference.  

To evaluate capability of handling scale variation of a network, we define {\bf density of connection} as $\mathcal{D}(\Delta  p)=\sum{\prod_{\Delta p}{w}}$ which means the sum of all paths between a neuron and its input at relative position $\Delta p$ with all $w$ set to 1. 
% A neural network usually consists of numerous connections that transfer information between neurons. 
There are usually numerous connections between neurons and each connection is assigned a weight $w$. A normal network learns the value of weights without changing the architecture of connections during training.
% better structure makes more reasonable density of connections
We prove in this part that GSL network we design is able to learn scales of each blocks which improves density of connections of the entire network, 
% we prove decomposition  makes density of connections unchanged
and that through scale decomposition our FD network could maintain a consistent density of connections with GSL network.
% while keeping fast and accessible to hardware optimization. 

\begin{figure}[t]
\begin{center}
    \includegraphics[width=8.2cm]{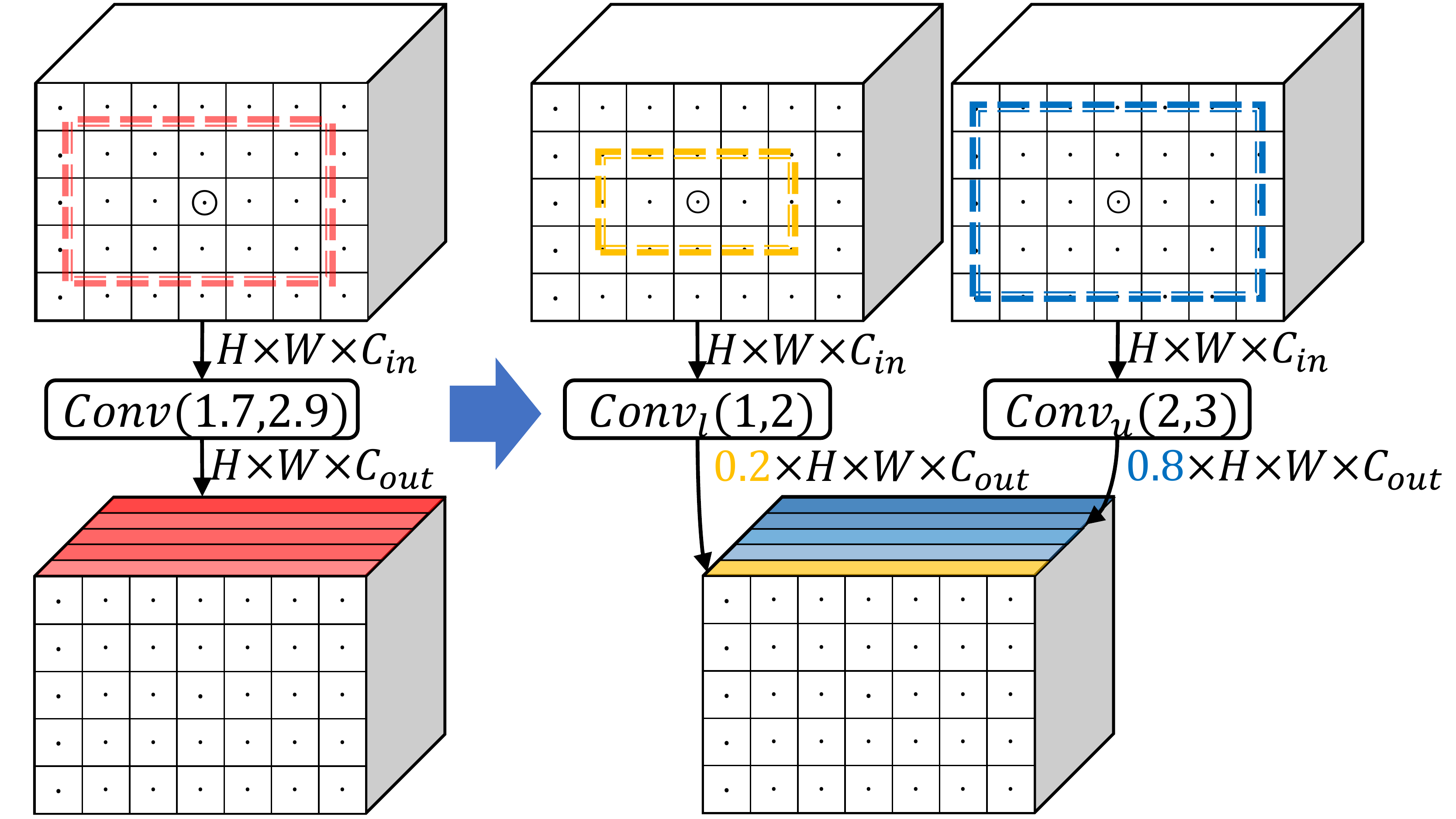}
\end{center}
\caption{
    An example of scale decomposition. A $3\times3$ convolution is split into a sub-conv with lower-bound dilations and a sub-conv with upper-bound dilations through scale decomposition.
}
\label{fig:decompose}
\end{figure}

\subsection{Global Scale Learner}
In our approach, we substitute 3$\times$3 convolutions with our global scale learning modules as shown in Fig~\ref{fig:pipeline} in GSL network. The scale learning module in our block is split into two parts, a global scale predictor and a convolution with bi-linear sampler which is widely used in methods like DCN\cite{dai2017deformable}, SAC\cite{zhang2017scale}, STN\cite{jaderberg2015spatial}. The global scale predictor takes the input feature map $U \in  \mathbb{R}^{H\times W \times C}$ with width $W$, height $H$, and $C$ channels, and feeds it into a global average pooling layer. A linear layer takes the pooled features of channels $C$ and outputs a two-channel global scales in both directions for each image.

Based on the predicted scale $(d_h, d_w)$, we sample the inputs and pass them to convolution. For example, $\mathcal{H}$ and $\mathcal{W}$ define relative position of inputs with respect to output in a 3$\times$3 convolution with dilations, input and output channels set to 1 for simplicity, which is,
\begin{equation*}
\mathcal{H} = \{ -1,0, 1 \}, \mathcal{W} = \{ -1,0, 1 \}.
\end{equation*}
Given kernel weight $w$ and input feature map $x$, the output feature at position($h_0$,$w_0$) could be obtained by: 
\begin{equation*}
y(h_0,w_0) = \sum_{i\in \mathcal{H}, j\in \mathcal{W}}{w_{ij}x(h_0+id_h, w_0+jd_w)}.
\end{equation*}

Since coordinates of $x$ at position $(h, w)=(h_0+id_h, w_0+jd_w)$ might be fractional, we sample value of input feature $x(h, w)$ through bi-linear interpolation, formulated as:

\begin{equation*}
x(h, w)= \sum_{h^*,w^*}{\lambda(h^*,h)\lambda(w^*, w)x(h^*, w^*)},
\end{equation*}
where $h^* \in \{ \left \lfloor h \right \rfloor,  \left \lceil h \right \rceil\}$
, $w^* \in \{ \left \lfloor w \right \rfloor,  \left \lceil w \right \rceil\}$ and
 $\lambda(p,q) = max(0, 1-|p-q|)$.

\vspace{4mm}
Taking ResNet with bottleneck~\cite{he2016deep} as an example, we apply our GSL module on all the $3\times3$ conv layers in stage {\em conv3}, {\em conv4}, and {\em conv5}. We also try to apply it on {\em conv2}, but the learnt scales are so close to 1 thus it is meaningless to learn global scale in {\em conv2}. 
% Explain robustness of learnt 
As mentioned in~\ref{sec:dist-dil}, the learnt scales are extremely stable across samples. 
% compare with dcn
The standard deviations of predicted global scales on the whole training set are less than $\mathbf{3\%}$ of the mean value, while standard deviations of local scales generated in deformable convolution is greater than $\mathbf{50\%}$ of the mean value(i.e. $5.3\pm3.3$ for small objects and $8.4\pm 4.5$ for large objects in res5c) as mentioned in \cite{dai2017deformable}.
Thus we randomly sample 500+ images in training set and are able to treat the average of predicted scales as fixed values. 
% \subsection{Distribution of Dynamic Global Dilation \bf{Moved to experiments???}}

\subsection{Scale Decomposition}
In this part, we propose a scale decomposition method that can transform fractional scales into combination into integral dilations for acceleration while keeping {\bf density of connections} consistent. Taking a 1-D convolution(kernel size=3) with a fractional scale rate $d$ as example, it takes input map with $C_{in}$ channels and output $C_{out}$ channels regardless of spatial size. For each location $p$ on the output feature map $\{y_j\}_{j \in C_{out}}$, we have
\begin{equation*}
 y_j(p) = \sum_{i=1}^{C_{in}}\sum_{\lambda \in \mathcal{L}}w^{\lambda}_{ij}x_i(p+\lambda d),
 \label{eq:g0}
\end{equation*}
where $\lambda$ denotes the relative location in $\mathcal{L} = \{-1, 0, 1\}$.

When $d$ is fractional, convolution with bilinear sampler would both take $x(p+\lambda \left \lfloor d \right \rfloor)$ and $x(p+\lambda \left \lceil d \right \rceil)$ as inputs with a split factor $\alpha$ for bilinear interpolation defined as $\alpha = d - \left \lfloor d \right \rfloor$. Here we have
\begin{equation*}
 y_j(p) = \sum_{i=1}^{C_{in}}\sum_{\lambda \in \mathcal{L}}w^{\lambda}_{ij}((1-\alpha)
 x_i(p+\lambda \left \lfloor d \right \rfloor) 
 + \alpha x_i(p+\lambda \left \lceil d \right \rceil)).
 \label{eq:g1}
\end{equation*}
And the density of connection at relative position $\lambda \left \lfloor d \right \rfloor$ and $\lambda \left \lceil d \right \rceil$ with respect to output maps $\{y_j\}_{j \in C_{out}}$ are  $(1-\alpha)C_{in}C_{out}$ and $\alpha C_{in}C_{out}$ respectively.

In this stage, a convolution with fractional scale is split into two sub-convolutions with dilation $\lambda \left \lfloor d \right \rfloor$ and $\lambda \left \lceil d \right \rceil$ respectively. The sub-conv with lower bound dilation $\lambda \left \lfloor d \right \rfloor$ takes $(1-\alpha)C_{out}$ output channels and the sub-conv with upper bound dilation $\lambda \left \lceil d \right \rceil$ takes the other $\alpha C_{out}$ output channels. Then we 
have output features $\{y_j^*\}_{j \in C_{out}}$ at position $p$ as:
\begin{align*}
 y_j^*(p) &= \sum_{i=1}^{C_{in}}\sum_{\lambda \in \mathcal{L}}w^{\lambda}_{ij}
 x_i(p+\lambda \left \lfloor d \right \rfloor), 
 \intertext{when $j \in [1, (1-\alpha) C_{out}]$, and}
        &= \sum_{i=1}^{C_{in}}\sum_{\lambda \in \mathcal{L}}w^{\lambda}_{ij}
  x_i(p+\lambda \left \lceil d \right \rceil)),
 \intertext{when $j \in ((1-\alpha) C_{out}, C_{out}].$}
 \label{eq:g1}
\end{align*}

\vspace{-10mm}
Through this scale decomposition, the density of connection at relative position $\lambda \left \lfloor d \right \rfloor$ and $\lambda \left \lceil d \right \rceil$ with respect to output maps $\{y_j^*\}_{j \in C_{out}}$ are $(1-\alpha)C_{in}C_{out}$ and $\alpha C_{in}C_{out}$ respectively, which are consistent with convolution with fractional scale. In particular, we learn scales in both $H$ and $W$ directions as $(d_h, d_w)$ and set the split factor $\alpha = (d_h - \left \lfloor d_h \right \rfloor + d_w - \left \lfloor d_w \right \rfloor)/2$ for simplicity\footnote{This simplification in 2-D case may change the density of connection a little, but does not matter. }. An illustration of decomposing a 2-D convolution with learnt scales $(1.7, 2.9)$ is in Figure~\ref{fig:decompose}. The split factor is computed as $\alpha = \frac{0.7+0.9}{2} = 0.8$ in this case.

In this way we transfer the interpolation in spatial domain to channel domain while keeping the density of connection between inputs and outputs almost unchanged. We reset dilations of  all $3\times3$ convolutions in conv\{3,4,5\} using the learnt global scales. For those convs whose learnt scales are fractional, they are replaced with two sub-convs inside as described above. Now that our model consists of only convolutions with fixed integral dilations, it can be easily optimized and be fast in practical scenarios. 

There are some special cases that the global scale learnt in $h$ or $w$ direction is less than 1, and the lower-bound dilation is set to 0 after scale decomposition. To handle this issue, we set the kernel size of corresponding direction to 1 which is equivalent to 0 dilation. Besides, if the learnt scale in one direction is very close to integer($\Delta \le$0.05), we do the decomposition on the other direction. For instance, if the learnt dilation is (2.02, 1.7), then the dilations of the two outcome convolutions are set to (2,1) and (2,2) respectively with $\alpha = 0.7$.

\subsection{Weight Transfer}

\begin{figure}[t]
\begin{center}
    \includegraphics[width=8.2cm]{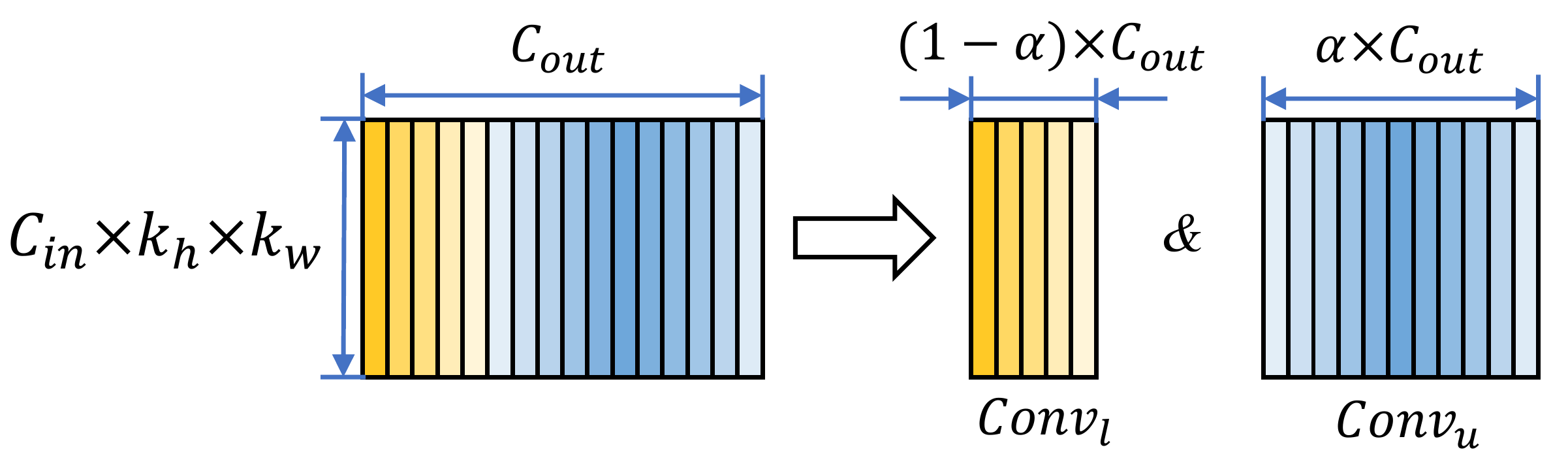}
\end{center}
\caption{
    Weight transfer of layers in FD network. When a normal convolution is split into two sub-convs, the convolution with lower-bound dilations takes the first $(1-\alpha)$ portion of pretrained weights in $C_{out}$ dim for initialization and the convolution with upper-bound dilations takes the last $\alpha$ portion of pretrained weights. 
}
\label{fig:weight-init}
\end{figure}

Since the density of connections in FD network is consistent with it in GSL network, it is free to take advantage of model weights in GSL network for initialization of FD network. 
Unlike conventional finetuning which directly copies weights of pretrained models, weights are also decomposed and transferred to layers in FD network in this paper.  
For convolutions assigned with integer scales, they are initialized with the weights of corresponding layers in pretrained model.
For convolutions with fractional scales that are split into two sub-convs with different dilations, the sub-conv with lower-bound dilations takes the weight of the first $(1-\alpha)$ output channels while the sub-conv with upper dilations takes the weight of the last $\alpha$ output channels from corresponding layer in pretrained model as demonstrated in Fig~\ref{fig:weight-init}.

\begin{figure}[t]
\centering
\begin{subfigure}[c]{0.15\textwidth}
\centering
\includegraphics[height=2.8cm]{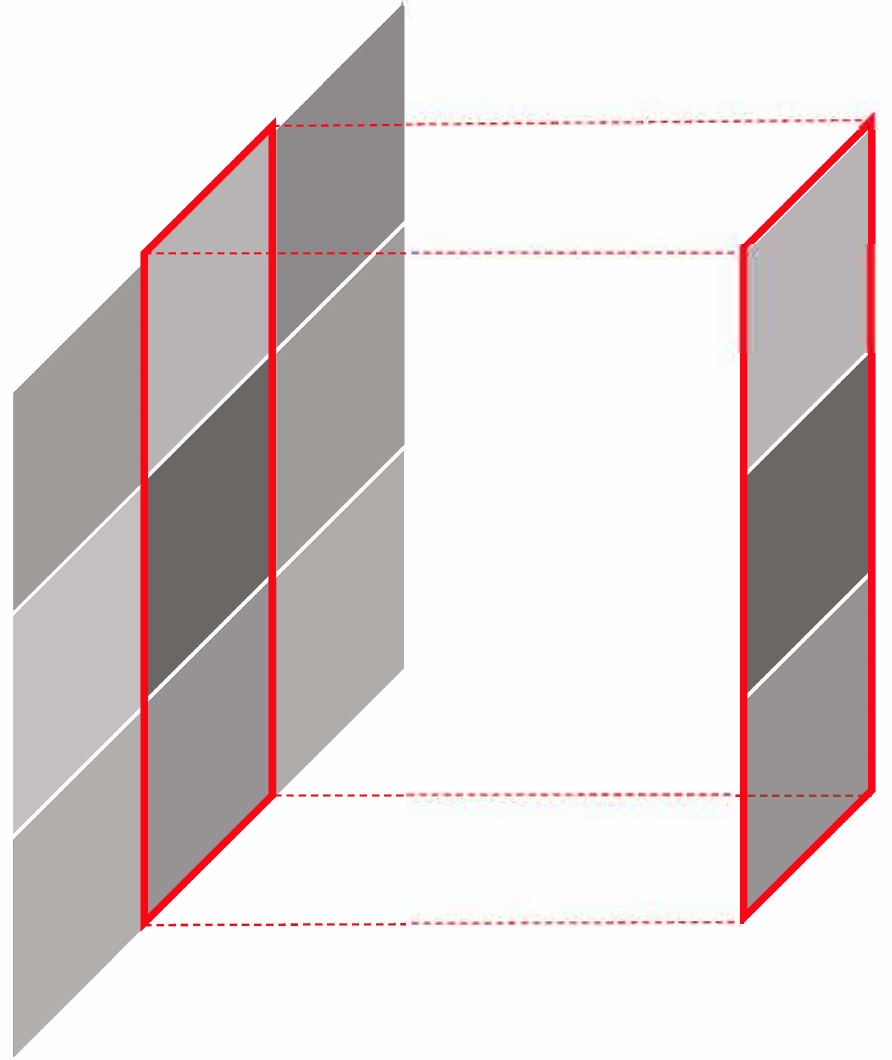}
\caption{3$\times$3 to 3$\times$1}
\end{subfigure}
\begin{subfigure}[c]{0.15\textwidth}
\centering
\includegraphics[height=2.8cm]{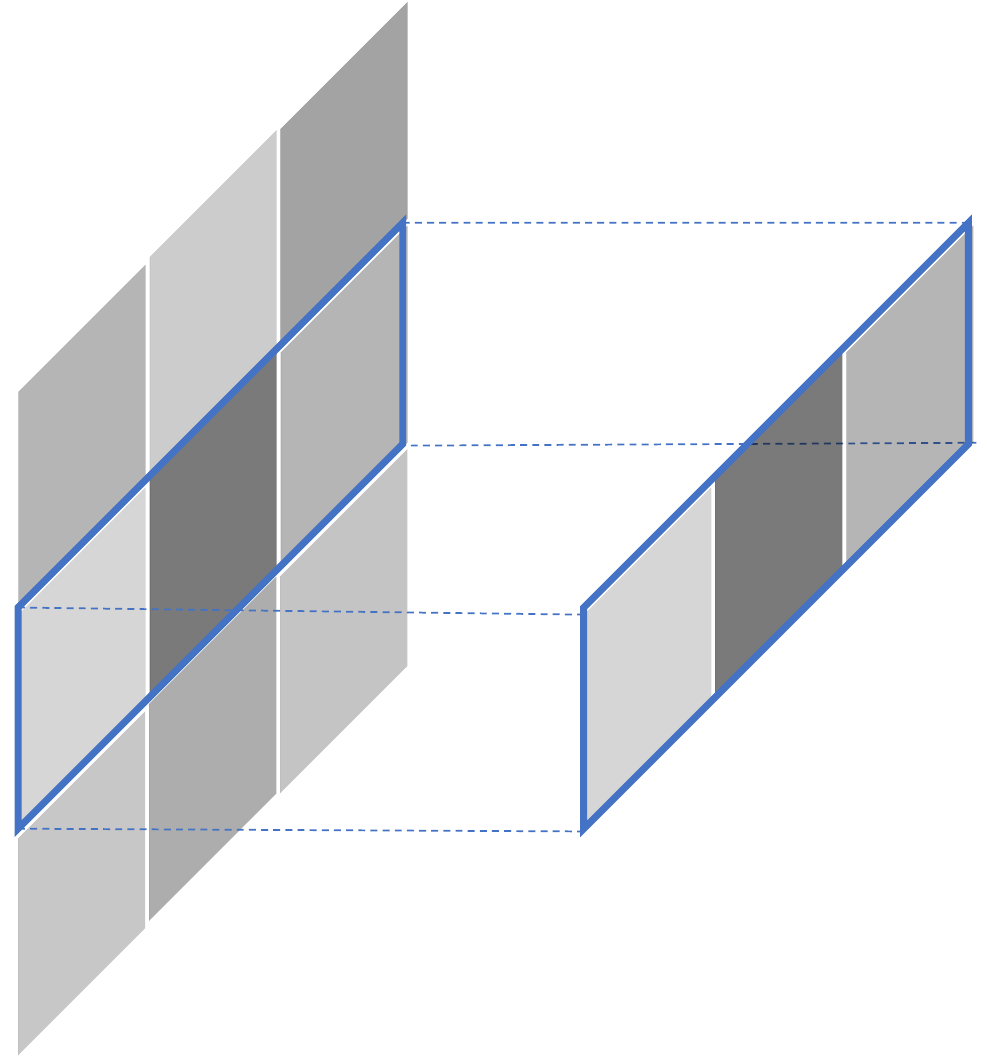}
\caption{3$\times$3 to 1$\times$3}
\end{subfigure}
\begin{subfigure}[c]{0.15\textwidth}
\centering
\includegraphics[height=2.8cm]{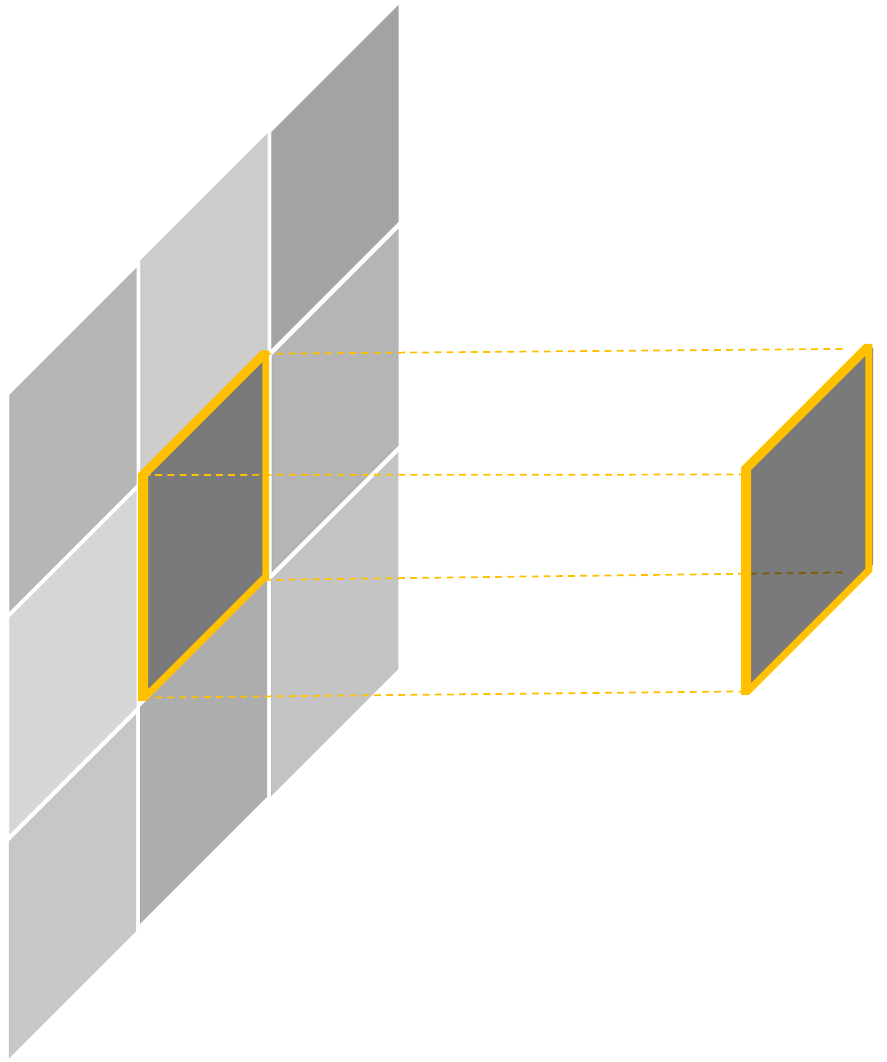}
\caption{3$\times$3 to 1$\times$1}
\end{subfigure}
\caption{Weight initialization in FD network for different cases of ($d_h, d_w$).}
\label{fig:weight-cases}
\end{figure}

In special cases that generates 1$\times$1, 1$\times$3 and 3$\times$1 sub-convs, weights are initialized with values of corresponding position in 3$\times$3 pretrained convolutions. Fig~\ref{fig:weight-cases} demonstrates how weights are initialized in different cases.

\begin{table*}[t]
	\centering %\addtolength{\tabcolsep}{-4pt}
	\footnotesize
	\setlength\tabcolsep{10pt}
	\begin{tabular}{c|c|ccc|ccc}
		Method & Backbone & AP & AP$_{50}$ & AP$_{75}$ & AP$_S$ & AP$_M$ & $AP_L$  \\
		\hline
		\hline
		YOLOv2 \cite{redmon2017yolo9000}             & DarkNet-19    &  21.6 & 44.0 & 19.2 & 5.0 & 22.4 & 35.5\\
		SSD-513 \cite{liu2016ssd}            & ResNet-101    &  31.2 & 50.4 & 33.3 & 10.2 & 34.5 & 49.8\\
		DSSD-513 \cite{fu2017dssd}           & ResNet-101    &  33.2 & 53.3 & 35.2 & 13.0 & 35.4 & 51.1\\
		RefineDet512 \cite{zhang2018single}       & ResNet-101    &  36.4 & 57.5 & 39.5 & 16.6 & 39.9 & 51.4\\
		RetinaNet \cite{lin2017focal}       & ResNet-101    &  39.1 & 59.1 & 42.3 & 21.8 & 42.7 & 50.2\\
% 		ConRetinaNet \cite{}       & ResNet-101    &  40.1 & 59.6 & 43.5 & 23.4 & 44.2 & 53.3\\
		CornerNet \cite{law2018cornernet}          & Hourglass-104 &  40.5 & 56.5 & 43.1 & 19.4 & 42.7 & 53.9\\
%  		RetinaNet800               & NASNet-A(6@4032)~\cite{NasNet2017iclr...} & 40.7 & - & - & - & - & - \\
	    \hline
	    \textbf{RetinaNet-FD-WT(ours)}     & ResNet-101     &  41.5 & 62.4 & 44.9 & 24.5 & 44.8 & 52.9\\
		\hline
		\hline
		FRCNN+++ \cite{ren2015faster}     & ResNet-101 &  34.9 & 55.7 & 37.4 & 15.6 & 38.7 & 50.9\\
		FRCNN w FPN \cite{lin2017feature}  & ResNet-101 &  36.2 & 59.1 & 39.0 & 18.2 & 39.0 & 48.2\\
		FRCNN w TDM \cite{shrivastava2016beyond}  & Inception-ResNet-v2 &  36.8 & 57.7 & 39.2 & 16.2 & 39.8 & 52.1\\
% 		D-FCN \cite{}  & Aligned-Inception-ResNet  &  37.5 & 58.0 & - & 19.4 & 40.1 & 52.5\\
		Regionlets \cite{xu2018deep}         & ResNet-101  &  39.3 & 59.8 & - & 21.7 & 43.7 & 50.9\\
		Mask R-CNN \cite{he2017mask}         & ResNet-101 &  38.2 & 60.3 & 41.7 & 20.1 & 41.1 & 50.2\\
		Fitness NMS \cite{tychsen2018improving}                  & ResNet-101 &  41.8 & 60.9 & 44.9 & 21.5 & 45.0 & 57.5 \\
		Cascade R-CNN \cite{cai2018cascade}  & ResNet-101 & 42.8 & 62.1 & 46.3 & 23.7 & 45.5 & 55.2 \\ 

        \hline
		\textbf{FRCNN-FD-WT(ours)}     & ResNet-101  &  42.1 & 63.4 & 45.7 & 21.8 & 45.1 & 57.1\\
        
	\end{tabular}\vspace{0.1cm}
	\caption{Comparison with state-of-the-art one-stage and two-stage detectors on COCO {\em test-dev}. }\label{tab:stoa}
	\vspace{-0.1in}

\end{table*}

%------------------------------------------------------------------------
\section{Experiments}
\label{sec:exps}
We present experimental results on the bounding box
detection track of the challenging MS COCO benchmark
~\cite{lin2014microsoft}. For training, we follow common practice~\cite{ren2015faster} and use
the MS COCO trainval35k split (union of 80k images
from train and a random 35k subset of images from the
40k image val split). If not specified, we report studies
by evaluating on the minival5k split. The COCO-style
Average Precision (AP) averages AP across IoU thresholds
from 0.5 to 0.95 with an interval of 0.05. These metrics
measure the detection performance of various qualities. Final results are also reported on the {\em test-dev} set.

\subsection{Training Details}
\label{sec:train-details}
We adopt the depth 50 or 101 ResNets w/o FPN as backbone of our model. In Faster-RCNN without FPN we adopt C4-backbone and C5-head while in Faster-RCNN with FPN and RetinaNet we adopt C5-backbone. RoI-align is chosen as our pooling strategy in both baselines and our models. 
We use SGD to optimize the training loss with 0.9 momentum and 0.0001 weight decay. 
The initial learning rate are set 0.00125 per image for both Faster-RCNN and RetinaNet. By convention, no data augmentations except standard horizontal flipping are used.
In our experiments, all models are trained following $1\times$ schedule, indicating that learning rate is divided by 10 at 8 and 11 epochs with a total of 13 epochs. 
We insists that it is responsible to compare different models with the same training schedule, so we list out training epochs in our experiments for fairness. Warming up and Synchronized BatchNorm mechanism~\cite{goyal2017accurate, peng2018megdet} are applied in both baselines and our method to make multi-GPU training more stable.

\begin{table*}[t]
	\centering %\addtolength{\tabcolsep}{-4pt}
	\footnotesize
	\setlength\tabcolsep{10pt}
	\begin{tabular}{c|c|c|ccc|ccc}
		Method &Type & Lr schd & AP & AP$_{50}$ & AP$_{75}$ & AP$_S$ & AP$_M$ & AP$_L$  \\
		\hline
		\hline
		FRCNN      & R50-C4 & $1\times$ & 35.4 & 55.6 & 37.9 & 18.1 & 39.4 & 50.3\\
% 		Faster-RCNN-FD   & R50-C4 & $1\times$ & 38.1 & 59.0 & 41.1 & 19.8 & 42.0 & 53.3\\
		FRCNN-FD   & R50-C4 & $1\times$ & \textbf{38.4(3.0)} & \textbf{59.3} & \textbf{41.5} & \textbf{19.8} & \textbf{42.3} & \textbf{55.3}\\
		\hline
		FRCNN      & R101-C4 & $1\times$ & 38.7 & 59.2 & 41.5 & 19.6 & 43.1 & 54.5\\
% 		Faster-RCNN-FD   & R101-C4 & $1\times$ & 40.2 & 61.5 & 43.1 & 21.3 & 44.7 & 56.2\\
		FRCNN-FD   & R101-C4 & $1\times$ & \textbf{40.9(2.2)} & \textbf{62.1} & \textbf{43.9} & \textbf{21.6} & \textbf{45.1} & \textbf{57.4}\\
		\hline
		\hline
		FRCNN      & R50-FPN & $1\times$ & 36.2 & 58.6 & 38.6 & 21.0 & 39.8 & 47.2\\
		FRCNN-FD   & R50-FPN & $1\times$ & \textbf{37.9(1.7)} & \textbf{60.6} & \textbf{40.8} & \textbf{22.3} & \textbf{41.2} & \textbf{49.6}\\
		\hline
		FRCNN      & R101-FPN & $1\times$ & 38.6 & 60.7 & 41.7 & 22.8 & 42.8 & 49.6\\
		FRCNN-FD   & R101-FPN & $1\times$ & \textbf{40.1(1.5)} & \textbf{62.8} & \textbf{43.3} & \textbf{23.4} & \textbf{44.6} & \textbf{52.3}\\
		\hline
		\hline
		Retina           & R50-FPN & $1\times$ & 36.0 & 56.1 & 38.6 & 20.4 & 40.0 & 48.2\\
		RetinaNet-FD        & R50-FPN & $1\times$ & \textbf{37.5(1.5)} & \textbf{57.7} & \textbf{40.2} & \textbf{20.7} & \textbf{41.0} & \textbf{50.3}\\
		\hline
		RetinaNet           & R101-FPN & $1\times$ & 37.5 & 57.2 & 40.1 & 20.5 & 41.8 & 50.2\\
		RetinaNet-FD        & R101-FPN & $1\times$ & \textbf{38.8(1.3)} & \textbf{59.0} & \textbf{41.7} & \textbf{21.6} & \textbf{42.9} & \textbf{52.6}\\
        
	\end{tabular}\vspace{0.1cm}
	\caption{Comparison with baselines on COCO {\em minival} using dilations decomposed from global scales learnt in GSL network. Our method is applied on different type of detectors with ResNet-50 and ResNet-101, and outperforms baselines in all cases. 
% 	* means that blocks in Res5-head are applied with our method. 
	All models are finetuned from ImageNet pretrained ResNets.} \label{tab:base}
	\vspace{-0.1in}

\end{table*}

In GSL network, parameters of backbone are initialized by ImageNet~\cite{russakovsky2015imagenet} pretrained model ~\cite{krizhevsky2012imagenet} straightforwardly, other new parameters are initialized by He (MSRA) initialization~\cite{he2015delving}. In FD network, parameters are initialized in the same way as GSL network if not specified. For FD models whose weights are transferred from GSL models, we add a suffix ``{\bf WT}'' to its name for clarity. 
% Name of FD models whose weights are transferred from GSL network would be added a suffix ``{\bf TF}'' for clarity.

\subsection{Ablation Study}
% \subsubsection*{Effectiveness of learnt dilations}
\vspace{0.15cm}
\noindent $\bullet$\;{\bf Effectiveness of Learnt Scales}

The effects of learnt scales are examined from ablations shown in Table~\ref{tab:base}. Faster-RCNN with C4-B+C5-Head, FPN and RetinaNet are included in this experiment, and ResNet-50 and ResNet-101 are adopted as backbone. In faster-RCNN using C4-backbone and C5-Head, we apply our method on {\em conv3} and {\em conv4} of backbones, and on {\em conv5} in head. Without bells and whistles, Faster-RCNN-FD with C4-backbone and C5-head on ResNet-50/101 could achieve 38.4/40.9 AP, which outperforms baselines by 3.0/2.2 points. As in Faster-RCNN-FD with FPN and RetinaNet-FD, using ResNet-50 as backbone yields 1.7 and 1.5 AP improvement respectively with neither extra parameters nor extra computational costs imported. Even in deeper backbone like ResNet101, the improvement is still considerable.
% (2.2 for FRCNN-R101-C4, 1.5 for FRCNN-R101-FPN and 1.3 for Retina-R101). 
Results in Table~\ref{tab:base} show that using combinations of fixed but more reasonable dilations without any other change could effectively improve the performance in tasks of detection. Note that Faster-RCNN-FD with C4-backbone and C5-head benefits a lot, while improvement on detectors with FPN are less but still remarkable.
% \textcolor{red}{Neither extra FLOPs nor parameters is imported in the process}.
\begin{table*}[h]
	\centering %\addtolength{\tabcolsep}{-4pt}
	\footnotesize
	\setlength\tabcolsep{6.1pt}
	\begin{tabular}{c|c|c|c|cc|cc|cc}
		Method & Dataset & Type & Scale Rate & $R3_1$ & $R3_4$ & $R4_1$ & $R4_6$ & $R5_1$ & $R5_3$ \\
		\hline
		\hline
		\multirow{2}{*}{FRCNN} & \multirow{2}{*}{{\em train}} & \multirow{2}{*}{R50-C4}  
		& h & $0.88\pm 0.00$ &  $1.30\pm0.01$ & $1.39\pm0.02$ & $8.49\pm0.28$ & $1.34\pm0.02$ & $3.57\pm0.02$ \\
		& & & w & $0.95\pm 0.00$ & $1.28\pm0.01$ & $1.41\pm0.02$ & $9.34\pm0.32$ & $1.20\pm0.01$ & $3.45\pm0.03$ \\
		\hline
		\multirow{2}{*}{FRCNN} & \multirow{2}{*}{{\em minival}} & \multirow{2}{*}{R50-C4}  
		& h & $0.88\pm 0.00$ &  $1.30\pm0.01$ & $1.39\pm0.02$ & $8.44\pm0.30$ & $1.34\pm0.02$ & $3.57\pm0.02$ \\
		& & & w & $0.95\pm 0.00$ & $1.28\pm0.01$ & $1.41\pm0.02$ & $9.31\pm0.31$ & $1.20\pm0.01$ & $3.45\pm0.03$ \\
		\hline
		\multirow{2}{*}{FRCNN} & \multirow{2}{*}{{\em train}} & \multirow{2}{*}{R50-FPN}  
		& h & $0.89\pm 0.00$ &  $1.19\pm0.01$ & $0.77\pm0.00$ & $1.41\pm0.01$ & $0.46\pm0.00$ & $2.81\pm0.04$ \\
		& & & w & $0.93\pm 0.00$ & $1.15\pm0.01$ & $0.82\pm0.00$ & $1.39\pm0.01$ & $0.89\pm0.00$ & $3.68\pm0.05$ \\
		\hline
		\multirow{2}{*}{RetinaNet} & \multirow{2}{*}{{\em train}} & \multirow{2}{*}{R50-FPN} 
		& h & $0.94\pm 0.00$ &  $1.19\pm0.00$ & $0.80\pm0.00$ & $1.46\pm0.02$ & $0.65\pm0.00$ & $1.75\pm0.03$ \\
		& & & w & $0.98\pm 0.00$ & $1.15\pm0.00$ & $0.85\pm0.00$ & $1.47\pm0.02$ & $0.91\pm0.00$ & $1.53\pm0.03$ \\

	\end{tabular}\vspace{0.1cm}
	\caption{Distribution of learnt scales for different blocks on different type of detectors or in different data splits.}\label{tab:dist-data}
	\vspace{-0.1in}

\end{table*}

% \subsubsection*{Dynamic-GD $vs.$ Fixed-GD}
\vspace{0.15cm}
\noindent $\bullet$\;{\bf GSL Network $vs.$ FD Network}

\begin{table}[!t]
\centering %\addtolength{\tabcolsep}{-1pt}
\footnotesize
\setlength\tabcolsep{6.7pt}
\begin{tabular}{c|c|c|c|c}
Settings & Lr schd & Type & AP & Inference time \\
\hline
\hline
FRCNN          & 1x & R50-C4 & 35.5 &  158ms\\
FRCNN-GSL   & 1x & R50-C4 & \textbf{38.7} &  269ms\\
FRCNN-FD   & 1x & R50-C4 & 38.4 &  159ms\\
\hline
\hline
FRCNN          & 1x & R101-C4 & 38.7 &  171ms\\
FRCNN-GSL   & 1x & R101-C4 & 40.7 &  281ms\\
FRCNN-FD   & 1x & R101-C4 & \textbf{40.9} &  172ms\\
\hline
\hline
RetinaNet            & 1x & R50-FPN  & 36.0 &  100ms\\
RetinaNet-GSL        & 1x & R50-FPN  & \textbf{38.0} &  150ms\\
RetinaNet-FD        & 1x & R50-FPN  & 37.5 &  101ms\\

\end{tabular}\vspace{0.1cm}
\caption{Comparison of performance and inference time between GSL network and FD network without any hardware optimization.}
\label{tab:speed}
\vspace{-0.1in}
\end{table}

Table~\ref{tab:speed} shows the performance and speed comparison between GSL network and FD network. Although GSL model can yield a considerable improvement on AP, it is slow and highly unfriendly to hardware acceleration due to unfixed indexing and interpolation operations. FD model keeps almost the same performance and can reach a relatively high speed even without any optimization. 

\vspace{0.15cm}
\noindent $\bullet$\;{\bf Weights Transferred from GSL Network}

We also investigate the effect of weight transfer from GSL model. The GSL models are trained with 1$\times$ lr schedule to learn global scales. For fairness, we finetune our FD model from GSL model using weight transfer within 1$\times$ lr schedule, and compare with baselines as well as FD models finetuned from ImageNet model with 2$\times$ lr schedule. Table~\ref{tab:finetune} shows the results of Faster-RCNN with R50-C4, R50-FPN structures and of RetinaNet with R50-FPN. Note that FD-1$\times$ finetuned from GSL model reaches a higher performance than FD-2$\times$ finetuned from ImageNet model and can outperform baseline-2$\times$ by a large margin in R50-C4 structure($3.0\%$AP) and Retina-R50-FPN($1.6\%$AP). As in Faster-RCNN with FPN, FD-1$\times$ model finetuned from GSL model can have a close performance to FD-2$\times$ model finetuned from ImageNet pretrained model. Thus, it is convenient to get a high performance model as 2$\times$ model with the same training epochs in total through this weight transfer strategy.

We also find model finetuned from ImageNet saturates in performance being trained for more than $2\times$ the epochs which is consistent to findings mentioned in \cite{he2018rethinking}.
As shown in Table~\ref{tab:finetune}, using weights transferred from GSL network could yield a further improvement on the saturated performance.

\begin{table}[!t]
\centering %\addtolength{\tabcolsep}{-1pt}
\footnotesize
\setlength\tabcolsep{6.7pt}
\begin{tabular}{c|c|c|c|p{0.15cm}p{0.15cm}p{0.25cm}}
Settings & Lr schd & Type & AP & AP$_{s}$ & AP$_{m}$ & AP$_{l}$ \\
\hline
\hline
FRCNN         & 2x & R50-C4 & 37.0 & 18.6 & 41.3 & 51.9 \\
FRCNN-FD      & 2x & R50-C4 & 39.6 & \textbf{20.7} & 43.6 & 56.5 \\
FRCNN-FD-WT   & 1x & R50-C4 & \textbf{40.0} & 20.6 & \textbf{44.1} & \textbf{56.8} \\
\hline
\hline
FRCNN         & 2x & R50-FPN & 37.5 & 22.1 & 40.8 & 48.4 \\
FRCNN-FD      & 2x & R50-FPN & \textbf{39.2} & \textbf{23.4} & \textbf{42.5} & 50.7 \\
FRCNN-FD-WT   & 1x & R50-FPN & 39.1 & 23.1 & \textbf{42.5} & \textbf{51.0} \\
\hline
\hline
RetinaNet         & 2x & R50-FPN & 36.7 & 19.9 & 40.4 & 48.4 \\
RetinaNet-FD      & 2x & R50-FPN & 37.8 & 21.0 & 41.0 & 50.7 \\
RetinaNet-FD-WT   & 1x & R50-FPN & \textbf{38.3} & \textbf{21.7} & \textbf{41.7} & \textbf{51.1} \\

\end{tabular}\vspace{0.1cm}
\caption{Comparison of models finetuned from $1\times$ GSL models with $1\times$ schedule and models trained with $2\times$ schedule. 
% Finetuning from GSL models could achieve a best performance in most cases while using a total $2\times$ training epochs as other models.
}
\label{tab:finetune}
\vspace{-0.1in}
\end{table}

\begin{figure*}[t]
\centering
\begin{subfigure}[c]{0.45\textwidth}
\centering
\includegraphics[height=4cm]{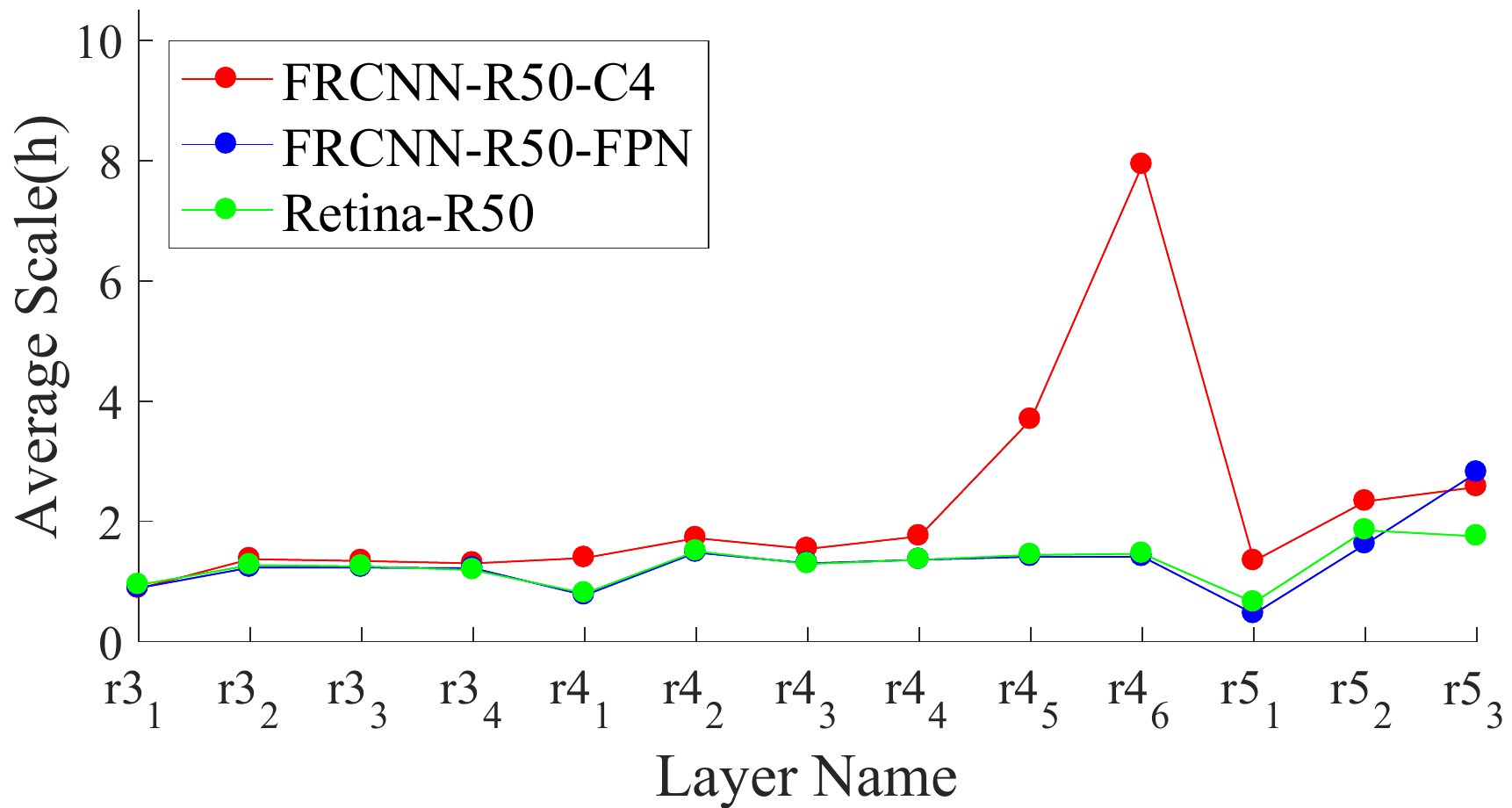}
\caption{Average global scales in $h$ direction.}
\end{subfigure}
\begin{subfigure}[c]{0.45\textwidth}
\centering
\includegraphics[height=4cm]{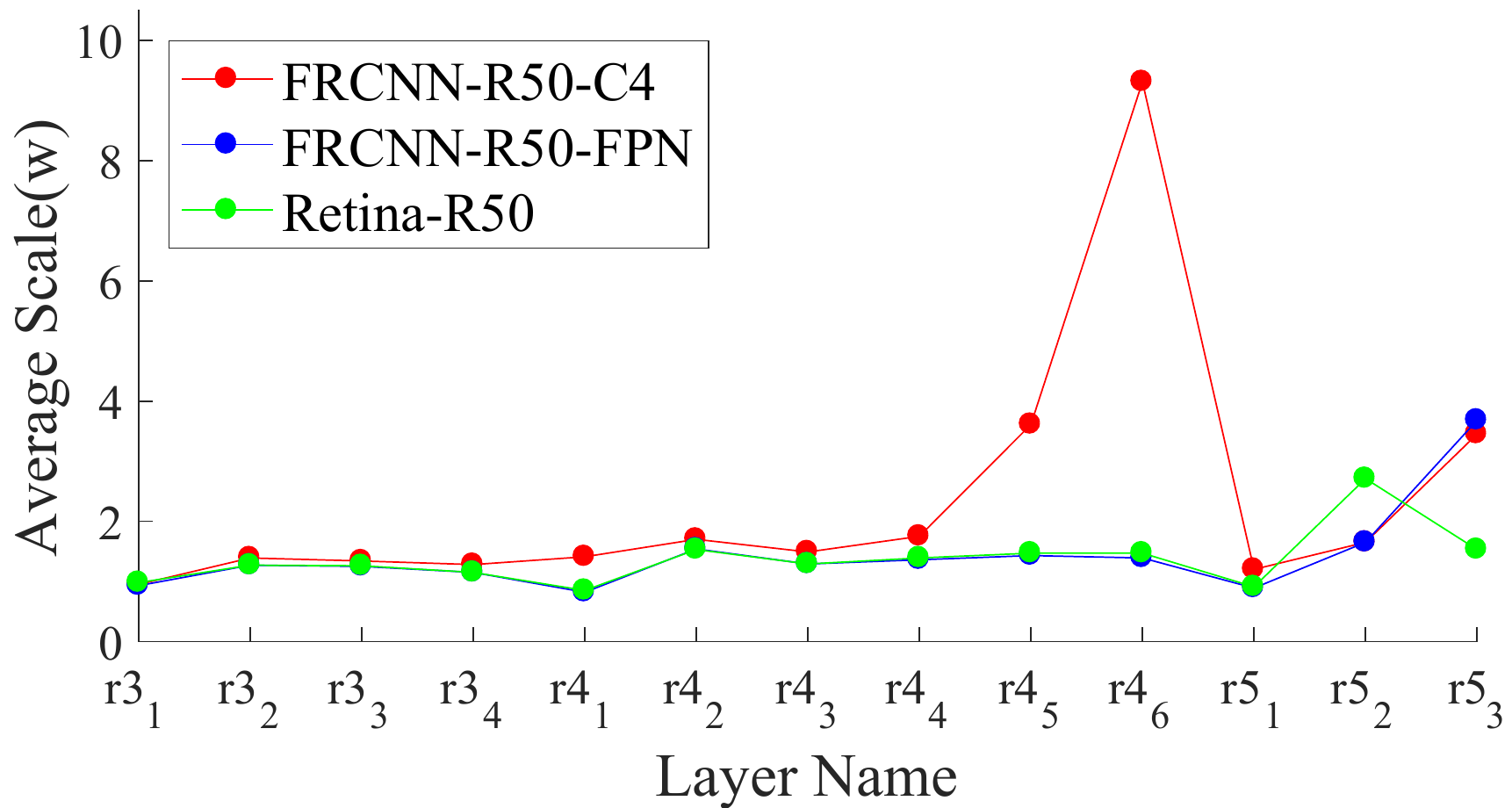}
\caption{Average global scales in $w$ direction.}
\end{subfigure}

\begin{subfigure}[c]{0.45\textwidth}
\centering
\includegraphics[height=4cm]{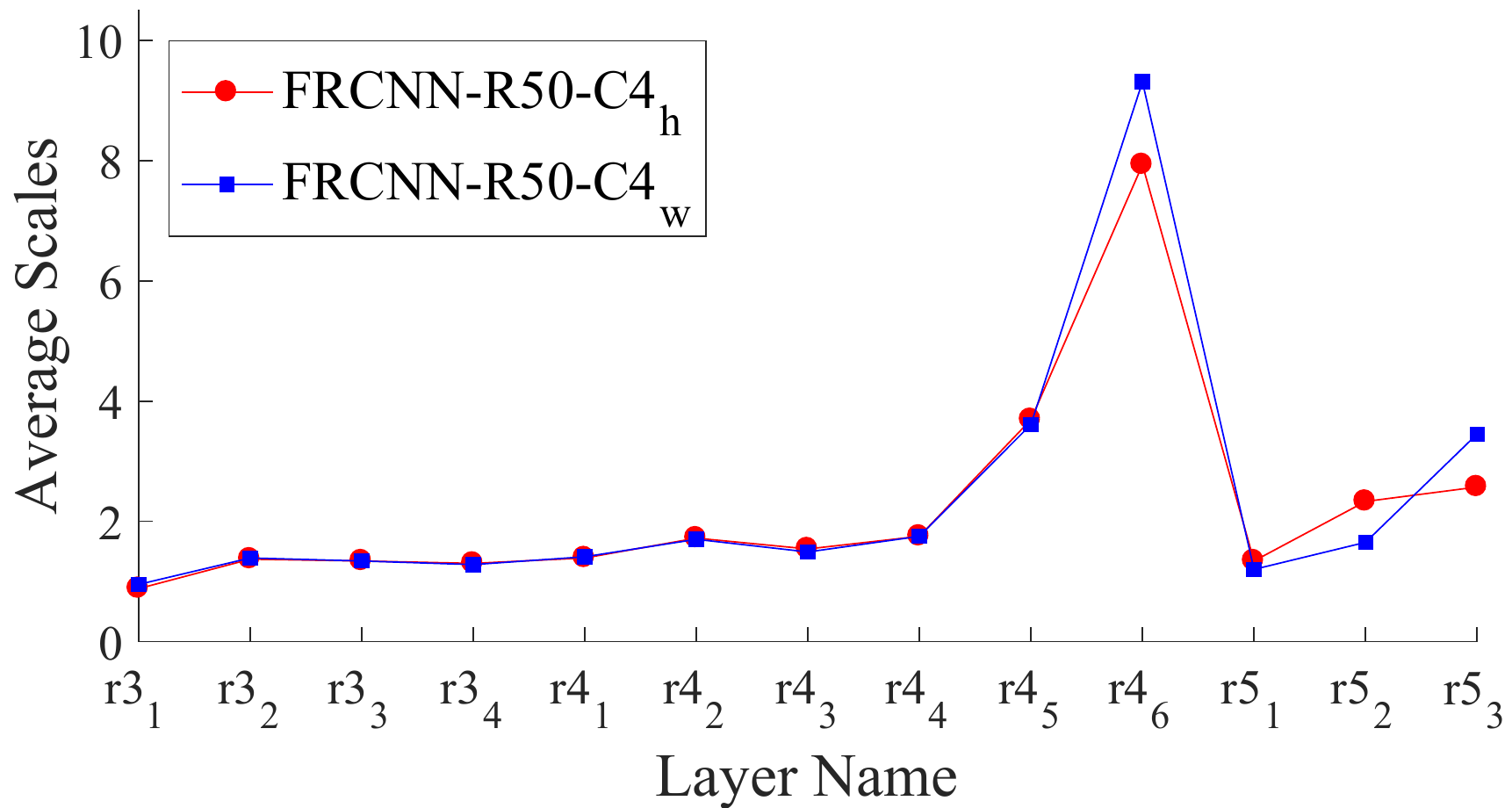}
\caption{Average global scales of Faster-RCNN.}
\end{subfigure}
\begin{subfigure}[c]{0.45\textwidth}
\centering
\includegraphics[height=4cm]{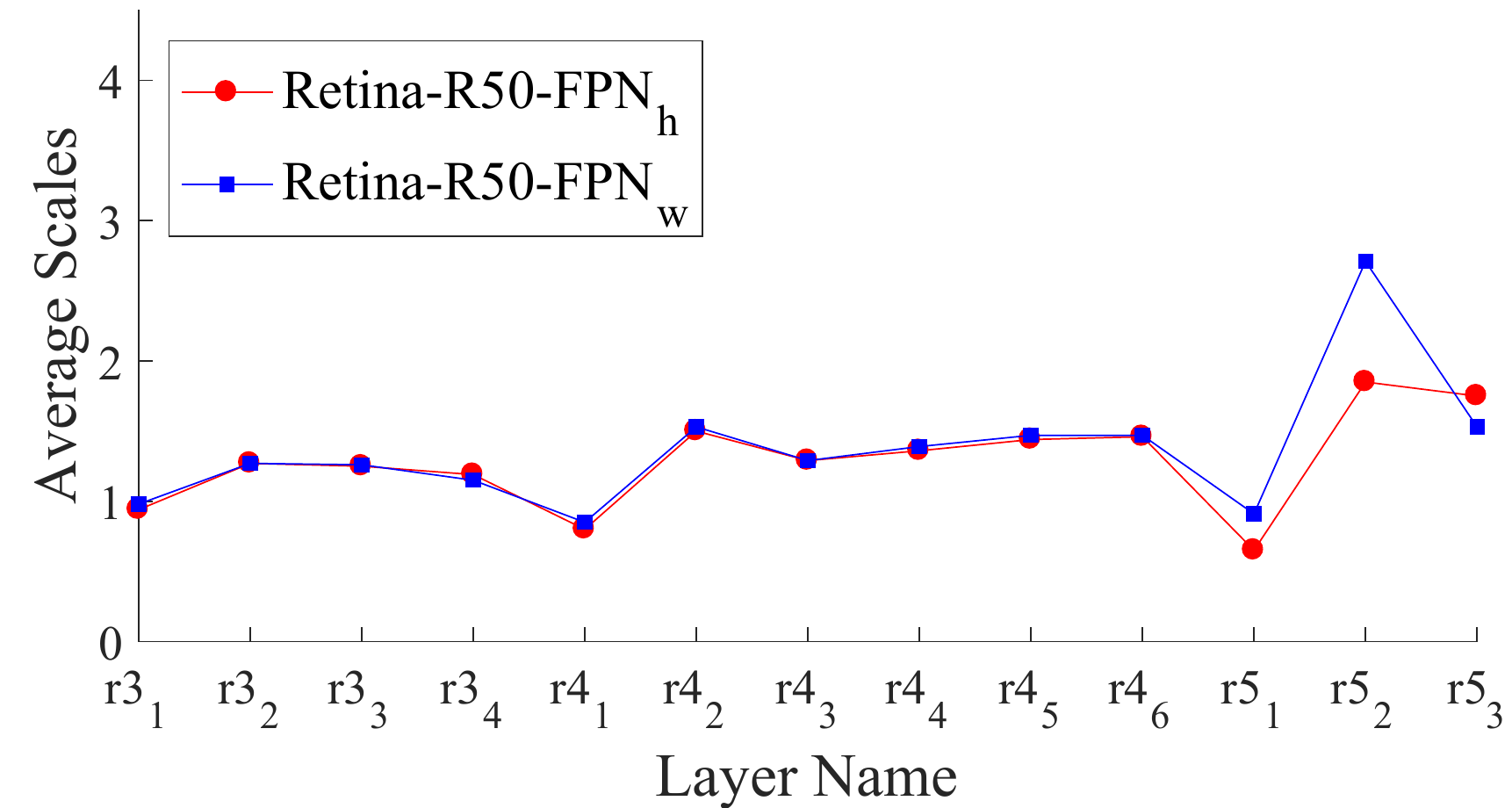}
\caption{Average global scales of RetinaNet.}
\end{subfigure}
\caption{Average of learnt global scales in different detection frameworks.}
\label{fig:dil_models}
\end{figure*}

\subsection{Compared with SOTA}
\label{sec:stoa}
For complete comparison, we evaluate our method on COCO \emph{test-dev} set. We adopt ResNet-101 in Faster-RCNN with C4-B+C5-Head and RetinaNet with FPN as our baselines. 
When compared with one-stage detectors, we apply exactly the same scale jitter as Retina800~\cite{lin2017focal} did, and are finetuned from GSL-1x model using 1.5$\times$ schedule. The total number of training epochs is $2.5\times$ the normal training epochs which is consistent with Retina800~\cite{lin2017focal}. 
While compared with two-stage detectors, we train our model in single scale and use 1$\times$ lr schedule without any whistles and bells for fair comparison. Weight of our models are initialized from GSL networks trained with $1\times$ lr schd which are free to use in our method. Results of both one-stage and two stages detectors are shown in Table~\ref{tab:stoa}. Applied on various frameworks, our FD-model based on ResNet-101-C4 and Retina-101-FPN could achieve 42.1
% \footnote{We provide the learnt scales of this model. Scale rates of Res3 are [1.05, 1.12, 1.20, 1.15], scale rates of Res4 are [1.16, 1.41, 1.54, 1.38, 1.30, 1.54, 1.45, 1.39, 1.40, 1.10, 1.35, 1.28, 1.57, 1.61, 1.82, 2.61, 5, 3.69, 3.84, 1.75, 1.29, [4, 7.7], [2.82, 2.57]], and scale rates of Res5 head are [1.25, [2.43, 1.63], 3.49].}  
and 41.5
% \footnote{We provide the learnt scales of this model. Scale rates of Res3 are [1.0, 1.11, 1.20, 1.13], scale rates of Res4 are [0.81, 1.33, 1.49, 1.34, 1.28, 1.47, 1.41, 1.37, 1.33, 1.10, 1.28, 1.24, 1.32, 1.45, 1.66, 1.52, 2.10, 1.40, 1.62, 1.45, 1.32, 1.48, 1.53], and scale rates of Res5 are [0.70, 2.72, 1.70].}  
AP respectively, which are competitive performances comparing with other state-of-the-art methods. More importantly, instead of introducing complex modules or extra parameters, our method just unlocks the potential of the original detectors themselves while keeping fast. 
Besides, this method could be easily applied to backbones like ResNext~\cite{xie2017aggregated} and mobileNetV2~\cite{sandler2018mobilenetv2} for better performance or speed. It could also be added into many kinds of frameworks like CornerNet~\cite{law2018cornernet}, and combined with various techniques like Cascade R-CNN~\cite{cai2018cascade}, thus there is still room for huge improvement on our method.

\subsection{Distribution of Learnt Scales}
\label{sec:dist-dil}

\indent To better understand our method, we think it is necessary to study to distribution of the learnt scales. Table~\ref{tab:dist-data} lists out the mean value of learnt scales of the first and last layers in each stages of different detectors based on ResNet-50. Observations are consistent in other layers. It is interesting to find in the first two rows of table that the learnt scales are very stable across samples and their average of each layer are almost identical in different data splits.
% that the standard deviations of predicted global dilations inside datasets are very small and average of them in different datasets are almost identical. 
This phenomenon demonstrates that it is reliable to use the average of global scales as configured fixed dilations in fast-deployment networks.  

Learnt scales of all layers are shown in Figure~\ref{fig:dil_models}. It is as expected that most layers with stride 1 require larger scale rates. There is also surprising findings that some layers like $res4_6$ in Faster-RCNN with R50-C4 structure demands dilations of $(8.49, 9.34)$ which are huge. Global scales of layers in model with FPN are smaller than those in model without FPN, this is because there are extra down-sampling stages like P5 and P6(even P7 in Retina) that alleviates the need of a very large receptive field in P4. Another interesting fact is that the learnt scales of convolutions with stride 2 in model with FPN are less than 1 without exception. One explanation is that enlarging receptive field by 2 times during each down-sampling stage is overmuch for FPN which needs to predict objects out from multiple stages, and a less-than-1 dilation can help shrink the receptive field size.  

\section{Conclusion}
% Propose
\vspace{-3mm}
We introduce a practical object detection method to improve scale-sensitivity of detectors. 
We design a GSL network to learn desirable global scales for different layers in detectors and they collaborate as unity to handle huge scale variation. Then the learnt global scales are decomposed into combination of integral dilations and a FD network is constructed and trained with the decomposed dilations and weights transferred from GSL network. 
The FD network enables detectors to be scale-sensitive and accurate while keeping fast and supportive of hardware acceleration. Extensive experiments show that our method is effective on various detection frameworks and very useful for practical applications.
% Future
Since our method is easy to be applied on various frameworks, we will try to combine it with those fastest frameworks to obtain most practical object detectors in future.
%

%------------------------------------------------------------------------
\section{Acknowledgements}
This work was supported in part by the National Key R\&D Program of China(No.2018YFB-1402605), the Beijing Municipal Natural Science Foundation (No.Z181100008918010), the National Natural Science Foundation of China(No.61836014, No.61761146004, No.61773375, No.61602481). The authors would like to thank NVAIL for their support.

{\small
\bibliographystyle{ieee_fullname}
\bibliography{egbib}
}

\end{document}